# Intelligent Road Anomaly Detection with Real-time Notification System for Enhanced Road Safety

Ali Almakhluk, Uthman Baroudi*, and Yasser El-Alfy
*Computer Engineering Department*
*Interdiscplnary Center for Intellgient Secure Systems*
*King Fahd University of Petroleum & Minerals*
Dhahran, Saudi Arabia
*Corresponding author: ubaroudi@kfupm.edu.sa

***Abstract— This study aims to improve transportation safety, especially traffic safety. Road damage anomalies such as potholes and cracks have emerged as a significant and recurring cause for accidents. To tackle this problem and improve road safety, a comprehensive system has been developed to detect potholes, cracks (e.g. alligator, transverse, longitudinal), classify their sizes, and transmit this data to the cloud for appropriate action by authorities. The system also broadcasts warning signals to nearby vehicles warning them if a severe anomaly is detected on the road. Moreover, the system can count road anomalies in real-time. It is emulated through the utilization of Raspberry Pi, a camera module, deep learning model, laptop, and cloud service. Deploying this innovative solution aims to proactively enhance road safety by notifying relevant authorities and drivers about the presence of potholes and cracks to take actions, thereby mitigating potential accidents arising from this prevalent road hazard leading to safer road conditions for the whole community.***

***Keywords— Edge computing, YOLO, Road Hazards, Real-time Notification System, Road Safety, Logistics.***

## I. Introduction

According to a 2018 report survey by the World Health Organization (WHO), 1.35 million people lose their lives in road accidents annually [5]. Enhancing the transportation sector health to improve transport safety, especially traffic safety, is a global objective that many countries worldwide are working on achieving by capitalizing on new technologies and disruptive trends in particular electric and autonomous vehicles, Artificial Intelligence (AI) and big data to achieve the above objectives [1]. Developing countries are particularly investing significant efforts into providing improved infrastructure for these road networks. Road accidents are closely linked to road conditions, and the presence of potholes and other anomalies can significantly impact driving style, leading to numerous undesirable situations [4].

The Internet of Things (IoT) is a burgeoning technology that facilitates communication between heterogenous computing and sensing devices. Presently, individuals are increasingly concerned about the condition of the roads they use. Consequently, it is important to monitor and maintain the surfaces of these roads in a novel manner [2]. Recently, a new technology called fog computing has emerged, which seeks to support low-latency services by moving the processing to devices at the network's edge, such as the roadside infrastructure or directly within connected vehicles [3]. The success of future Intelligent Transportation Systems (ITS) heavily relies on the deployment of sensors and actuators on vehicles and alongside the road. This infrastructure for sensing and actuation will enable the development of new applications aimed at enhancing various aspects of transportation, including safety through the distribution of road hazard warnings and traffic efficiency via the timely detection of traffic congestion [3]. The rapid expansion of the transportation industry has resulted in a corresponding growth in road networks.

This study is an effort to realize the objectives of NTS 2030 in Saudi Arabia [1] and other counties in the World as well. It aims to design a system that mitigates vehicle accidents caused by road potholes by utilizing low-cost, portable, and fast components. The proposed system comprises end-nodes, each representing a vehicle and equipped with a camera to capture visual data with a deep learning model to detect road hazards such as potholes and cracks and analyze their sizes, and a Roadside Unit (RSU) to collect and forward data from and to other end-nodes. Furthermore, the RSU will upload the images of road hazards to the cloud to notify the relevant authorities. Also, the RSU will transmit warning messages to other vehicles based on the size of the detected anomaly. Lastly, the system counts the anomalies on roads' surfaces based on a factor calculated using vehicle speed and frames captured per second.

The rest of the paper is organized as follows. In section II, we briefly present the existing works in the literature. Then, in section III, we explain the methodology in detail. In section IV, experiments and results are presented and discussed. Section V suggests further extraction to the work. Section VI briefly concludes what has been done in the work. Lastly, section VII shows all the references used in this work.

## II. Literature review

In this section, we will discuss the existing work done in this field. Based on [6][7][8], there are three common pothole detection methods, namely vibrational-based methods, vision-based methods, and 3D reconstruction-based methods. Table 1 compares the three methods for different criteria.

Table 1: Comparison among the three detection methods.

| Criterion | Vibrational based | Vision-based | 3D reconstruction-based |
|---|---|---|---|
| Processing time | Fastest | Middle | Slowest |
| Cost | Cheap | Middle | Expensive |
| Storage (data size) | Small | Middle | Large |
| Accuracy | Lowest | Middle | Highest |
| Can provide exact shape of pothole | No | No | Yes |
| Data type | Numerical data from accelerometer, ultrasonic sensor, or gyroscope | 2D images or videos | Stereo vision technology, laser reconstruction or Microsoft Kinect |
| Data processing could be done in low-cost devices | Yes | Yes | No |

There are many potholes detection deep learning models trained using different algorithms. Some were trained using ResNet-2, ResNet-18 and VGG-11 such as [4] with a combination of public and private datasets that contains 3211 pothole images classified as small, large and normal roads. The work in [9] used a combination of 4 public datasets that contains potholes, Bumps, Cracks, and no anomaly. Others used YOLO from YOLOv1 to YOLOv5 such as [10][11]. For [10], they used the COCO dataset that contains 5000 images of potholes, vehicles, animals and pedestrians and they tested the system in real world to evaluate the accuracy and response time and they mentioned that the system can accurately detect obstacles within a distance around 56m. Although they were able to detect obstacles that were further than 56m, they could not find the distance between them. Also, they were able to plot the markers with an update rate of 1.8 to 2 seconds. Furthermore, they were able to successfully generate phone calls without being cut off and SMS messages were sent within seconds. While in [11], they used one public dataset that contains 665 images for potholes and used Precision, Recall, F1-score, mAP@0.5, and interference time to evaluate their work. Some used YOLOX such as [12] and they utilized the same dataset in [11] that contains 665 pothole images and they evaluated their model using Epoch, mAP, Size, and interference time KPIs. Furthermore, a few papers introduced fully functioning systems using vibration-based methods for pothole detection such as [8] but they did not implement or test a prototype. Moreover, some papers introduced pothole detection systems using 3D-reconstruction-based methods such as [10][13]. In [13], they used a private dataset generated from a survey vehicle in Korea that contains 150 video clips, and they evaluated their work by choosing 90 random images. Few papers introduced and implemented a prototype using vision-based methods as in [14]. Furthermore, most papers classify potholes as pothole or non-pothole, and few differentiate between road anomalies like cracks or potholes but very few studies differentiate between road anomalies based on their size using vison-based methods such as [4].

This project will focus on developing an edge-based system for anomaly detection and classification using vision-based methods to provide instant vehicle warnings leveraging edge computing capabilities. Moreover, this project will evaluate the performance of the recently released YOLOv8 algorithm for training the model. Based on [15][16][17][18], YOLOv8 showed better performance in terms of accuracy and interference time compared to previous YOLO versions. The results will be compared against models trained with the RDD2022 dataset used in CRDDC 2022 [19]. Table 2 shows related works with the datasets and KPIs used to measure the performance.

Table 2: Public datasets and KPIs used to measure their performance.

| Year | Ref | Dataset summary | KPIs |
|---|---|---|---|
| 2023 | [4] | A combination of public and private datasets that contain 3211 images of potholes classified as small, large, and normal roads | Precision, Recall, F1- score |
| 2021 | [9] | Multiple public datasets were used for 4 road anomalies (pothole, Bump, Crack, and no Anomaly). | F1-Score, Accuracy, Recall, Precision. |
| 2022 | [10] | COCO dataset that contains 5000 images and they broke down the classes into 4 categories (pothole, vehicle, animal, and pedestrian). | Real world test; accuracy, response time, SMS messages, and phone calls without cut off. |
| 2022 | [11] | One public dataset that contains 665 pothole images. | Precision, Recall, F1-score, mAP@0.5, and interference time. |
| 2022 | [12] | Public dataset that contains 665 pothole images annotated | Epoch, mAP, Size, interference time |
| 2015 | [13] | Private dataset generated from a survey vehicle in Korea that contains 150 video clips. Use only 90 images selected randomly. | Accuracy, Precision, and recall. |

## III. Methodology

The methodology employed in this system involves an end node that represents a surveillance car that scans road surface and collects visual data using a 2D camera and processes them to detect road anomalies such as potholes and cracks. Then, it classifies each type based on its features (e.g., size). The system is incorporating the YOLOv8 model to detect road anomalies. It also classifies the anomalies into four categories (Potholes, Longitudinal Cracks, Transverse Cracks, and Alligator Cracks). Furthermore, if any relatively large anomaly is detected a warning signal will be broadcasted to all other nearby vehicles through the Roadside Unit (RSU). Also, for each road anomaly detected from potholes or cracks, an image of that anomaly will be uploaded to the cloud through Google Drive in order for the responsible authorities to take the proper action.

Moreover, the developed system has a Graphical User Interface (GUI) that controls the system operation modes and visualizes the analyzed footage shown in Figure 1. The GUI shows a counter for different classifications and interactable buttons to control the system operations to start and stop the analysis and switch between using prerecorded footage and live footage. In addition, the system will provide the user with an option to adjust the frame rate of the footage to calibrate it with the surveillance car. A small-scale prototype is built to emulate the system's functionalities. The following subsections explain the system modes of operations.

### A. *Mode-I: Visual data processed at the car side*

In this mode of operation, the camera is connected to the Raspberry Pi 4B to collect the visual data and do the processing on the Raspberry Pi by inferencing the trained deep learning model on it. Then, if any anomaly is detected, the video frame of the detected anomaly will be sent to the RSU (in this case, the laptop) through Wi-Fi using a server-client approach. The RSU will broadcast a signal to all other nearby vehicles connected to it. Also, if the number of detected anomalies exceeds a pre-specified threshold, a general signal will be broadcasted to warn vehicles that the road contains a lot of anomalies. Furthermore, RSU will upload all images that contain road anomalies to the cloud. Figure 2 illustrates the operation of Mode-I.

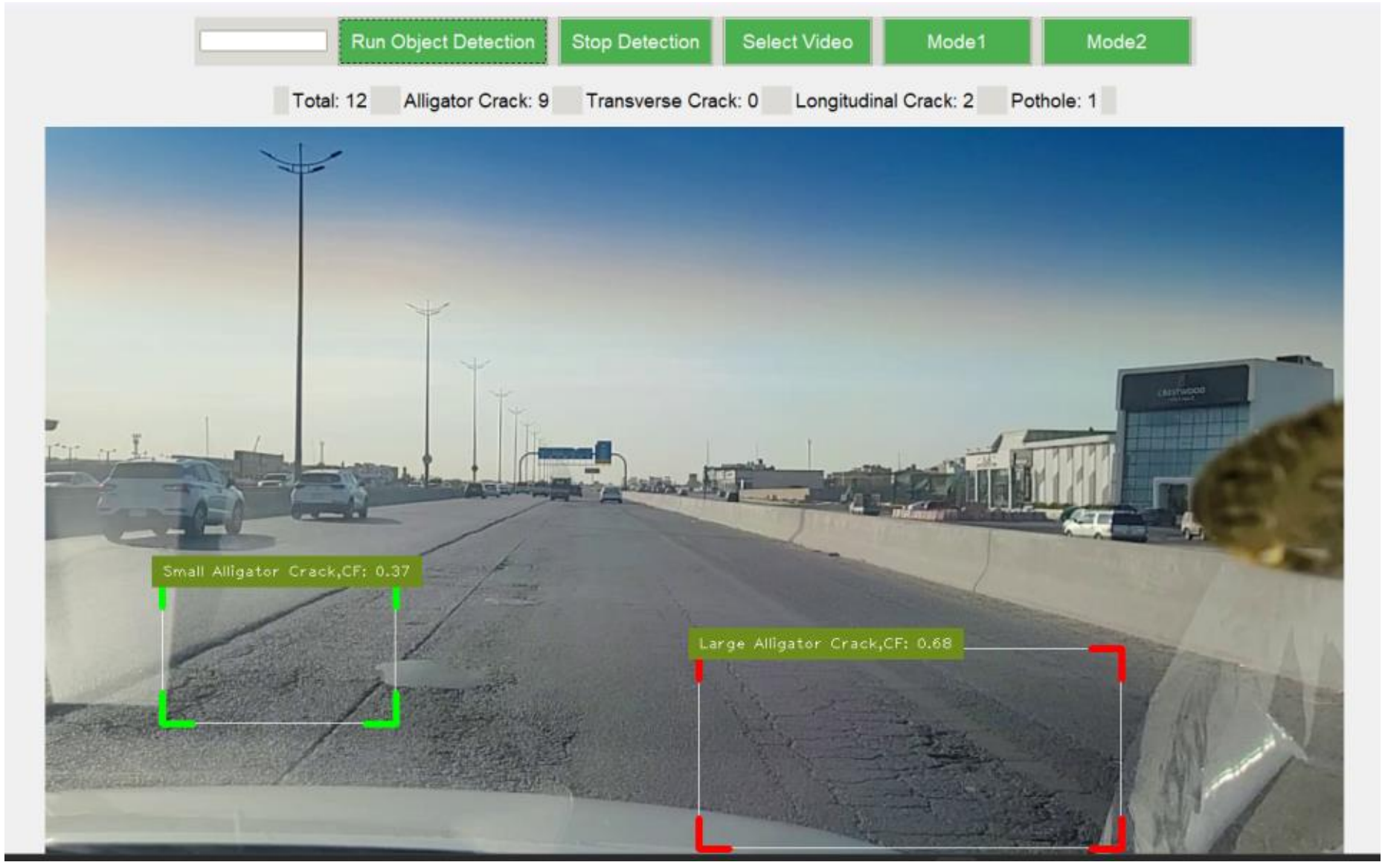

*Figure 1: GUI for the developed system.*

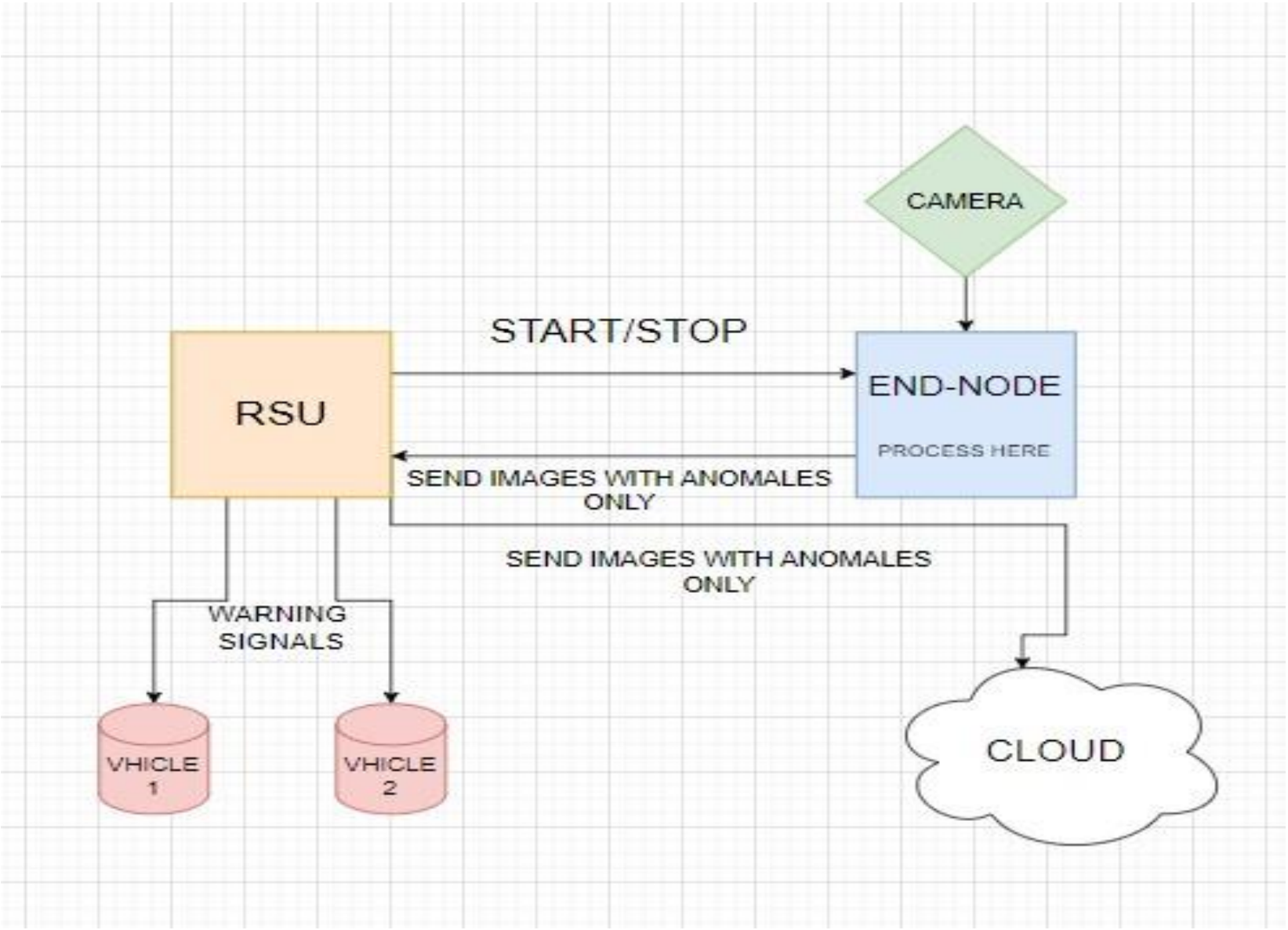

Figure 2: The system schematic diagram of Mode-I.

### B. *Mode-II: Visual data processed at the RSU side*

In this mode of operation, the visual data is collected by the camera connected to Raspberry Pi 4B, but it will be streamed to the RSU (in this case, the laptop) using UDP protocol to process the collected data. The motivation for this mode is the abundance of energy and computing power in RSUs. Then, if an anomaly is detected, a signal will be broadcasted to nearby vehicles via Wi-Fi. As in Mode-I, if the number of detected anomalies exceeds a pre-specified threshold, a general signal will be broadcasted to warn vehicles that the road contains a lot of anomalies. Furthermore, the RSU will upload all images that contain road anomalies to the cloud. Figure 3 illustrates the operation of Mode-II.

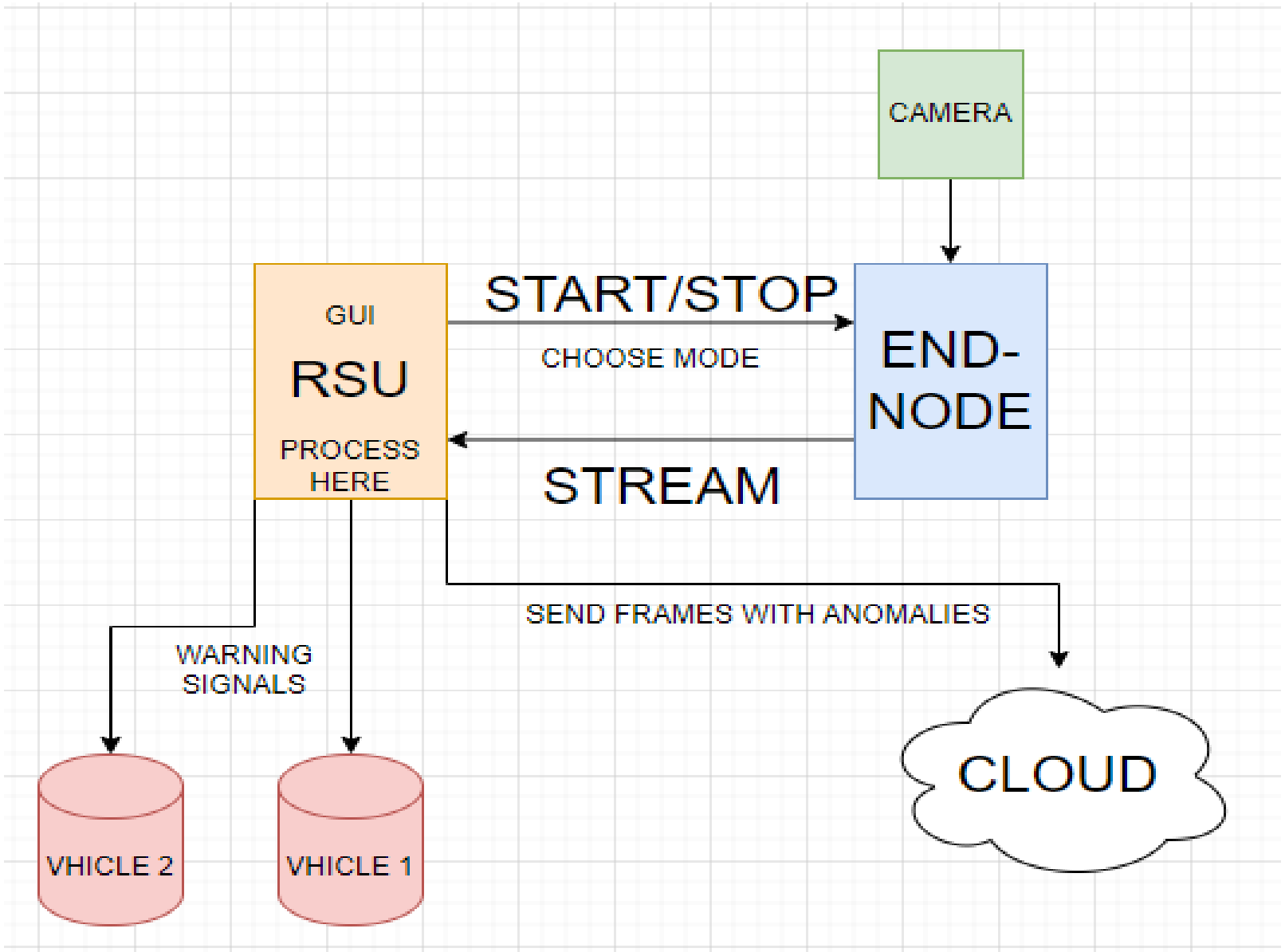

Figure 3: The system schematic diagram of Mode-II.

### C. *Anomaly Assessment*

1. Anomaly size estimation:

   To estimate the size of the detected anomaly, we use the Bounding Anomaly Box Area ($A$) as in Figure 1. $A = L.W$, where $L$ is the Bounding Box Length ($L$) and $W$ is the Bounding Box Width ($W$).

$$if \begin{cases} A \geq \rho, large\ anomaly \\ Otherwise, small\ anomaly \end{cases} \tag{1}$$

   where $\rho$ is the pre-set threshold in terms of pixels, which constitutes the anomaly size with respect to the whole image size. In our study, we set $\rho = 10\%$. For example, if the image size is 640 * 640, then $\rho = 40960$.

2. Classification models quality:

   To measure the classification model's quality, we use the following metrics:

$$\text{Precision} = \frac{(TP)}{(TP + FP)} \tag{2}$$

$$\text{Recalls} = \frac{(TP)}{(TP+FN)} \tag{3}$$

$$\text{F1-Score} = 2\times \frac{(Precision \times Recall)}{(Precision + Recall)} \tag{4}$$

3. Counting anomalies:

   Using life video frames for counting the detected anomalies poses several challenges. First, it is time and energy consuming. Second, it is misleading as the same anomaly is counted several times. Therefore, we propose the following approach to skip redundant video frames such that it counts the detected anomaly in a specific spot just once. We define a new term called Frame Span Interval (FSI), which is defined as follows:

$$FSI = \frac{Vechicle\ speed\ (m/s)}{fps} \quad \text{(m/frame)} \tag{5}$$

   where *fps* is a camera setting of the number of frames per second. Therefore, when the vehicle is moving at high speed, each frame spans a large space and hence it might observe multiple anomalies. In contrast, when the vehicle is moving at low speed, each frame spans a small space and hence it might observe the same anomaly over multiple frames. Later, in the developed system, we adopted the following rule:

$$if \begin{cases} FSI \geq 0.5, & \text{skip 5 frames} \\ Otherwise, & \text{skip 30 frames} \end{cases} \tag{6}$$

## D. *Proposed Modes Assessment*

1. Inference time:

   It is the time required for the CNN model to decide whether there is an identified object.

2. Storage requirements:

   Considering the adopted modes of operations, it is important to evaluate the required storage either in RSUs or vehicles.

# IV. RESULTS AND DISCUSSION

In this section, we will discuss in detail the system framework, experimental setup, obtained results, and analysis of obtained results.

## A) Framework

### *a) Models background*

There are multiple pre-trained CNNs (Convolutional Neural Networks) **Y**ou **O**nly **L**ook **O**nce version 8 (YOLOv8) models by Ultralytics, which are YOLOv8n, YOLOv8s, YOLOv8m, YOLOv8l and YOLOv8x [18]. Similarly, for YOLOv5 there are multiple ones such as YOLOv5n,

YOLOv5s, YOLOv5m, YOLOv5l, YOLOv5x [20]. Since the large ones (YOLOv5l, YOLOv5x, YOLOv8l and YOLOv8x) need high processing power, we train the first 3 n, s and m for both YOLOv5 and YOLOv8 to compare the results among the models, but the model that will be used in our final system will be YOLOv8.

*b) Dataset background*

We trained our model using the RDD2022 dataset [21], which contains 47420 images divided into 9035 test images without labels and 38385 training images with 55007 labels for 8 categories D00 (Longitudinal crack), D10 (Transverse crack), D20 (Alligator crack), D40 (Pothole), D43 (Crosswalk blur), D44 (While line blur), D50 (Manhole cover). This dataset was collected from six different countries: Japan, India, Czech, Norway, United states, and China. The image capturing devices are different such as smartphones, high-resolution cameras, Google Street view images using different vehicles such as cars, motorbikes, and drones [22].

*a) Dataset Pre-processing*

Since the RDD2022 dataset comes with XML annotation files, while YOLO supports a different annotation file type, which is txt files, Roboflow [17] has been used to do the conversion step. Furthermore, the images were split into 80:20 for training and validation, respectively. Then the first model was trained on the pre-trained model YOLOv8s for the 8 classes. additionally, since the results provided in [23] using the CRDDC2022 dataset selected only four classes: D00, D10, D20, and D40, we have dropped the other classes to have a fair comparison with contemporary approaches.

*b) Training the models*

The models (YOLOv5n, YOLOv5s, YOLOv5m, YOLOv8n, YOLOv8s and YOLOv8m for 4 classes D00, D10, D20 and D40) and (YOLOv8n and YOLOv8s for 8 classes D00, D10, D20, D40, D43, D44, D50 and repairs) are trained offline with Pytorch environment using a workstation empowered by Intel core I9 13700K CPU at 5 GHz, NVIDIA RTX3090Ti GPU and 64 GB RAM. All models were trained for 300 epochs, a batch size of 16, an image size of 640, and a patience set to 100.

*c) Results of trained models*

We used pre-trained weights from [20] for YOLOv5 and [18] for YOLOv8 for training our models on the custom dataset [21]. The fastest model to train was YOLOv8s (4 classes) which took 10.858 hours for 187 epochs with a size of 22.5MB, mAP50 = 0.597, mAP50-95 = 0.315, F1-score = 0.59, inference time while running it on raspberry pi 4 (S1) = 3.054 seconds per frame and inference time while running it on the laptop (S2) = 0.376 seconds per frame. The second fastest to train was YOLOv8s (8 classes) which took 12.423 hours for 200 epochs with a size of 22.5MB, mAP50 = 0.707, mAP50-95 = 0.437, F1-score = 0.67, inference time while running it on raspberry pi 4 (S1) = 3.054 seconds per frame and inference time while running it on the laptop (S2) = 0.376 seconds per frame. The third fastest to train was YOLOv8n (4 classes) which took 12.987 hours for 274 epochs with a size of 6.2MB, mAP50 = 0.577, mAP50-95 = 0.3, F1-score = 0.58, inference time while running it on raspberry pi 4 (S1) = 1.035 seconds per frame and inference time while running it on the laptop (S2) = 0.172 seconds per frame. The fourth fastest to train was YOLOv5n (4 classes) which took 14.783 hours for 300 epochs with a size of 3.9MB, mAP50 = 0.54, mAP50-95 = 0.249, F1-score = 0.56, inference time while running it on raspberry pi 4 (S1) = 1.074 seconds per frame and inference time while running it on the laptop (S2) = 0.231 seconds per frame. The fifth fastest to train was YOLOv5s (4 classes) which took 14.847 hours for 262 epochs with a size of 14.5MB, mAP50 = 0.568, mAP50-95 = 0.272, F1-score = 0.58, inference time while running it on raspberry pi 4 (S1) = 3.112 seconds per frame and inference time while running it on the laptop (S2) = 0.412 seconds per frame. The sixth fastest to train was YOLOv8n (8 classes) for 296 epochs with a size of 6.2MB, mAP50 = 0.689, mAP50-95 = 0.423, F1-score = 0.66, inference time while running it on raspberry pi 4 (S1) = 1.035 seconds per frame and inference time while running it on the laptop (S2) = 0.172 seconds per frame. The seventh fastest was YOLOv8m (4 classes) which took 15.466 hours for 160 epochs with a size of 52.0MB, mAP50 = 0.615, mAP50-95 = 0.326, F1-score = 0.60, inference time while running it on raspberry pi 4 (S1) = 4.717 seconds per frame and inference time while running it on the laptop (S2) = 0.737 seconds per frame. The eighth fastest was YOLOv5m (4 classes) which took 16.386 hours for 198 epochs with a size of 42.2MB, mAP50 = 0.597, mAP50-95 = 0.297, F1-score = 0.60, inference time while running it on raspberry pi 4 (S1) = 4.748 seconds per frame and inference time while running it on the laptop (S2) = 0.778 seconds per frame. Lastly, we have YOLOv8x (8 classes) which took 32.526 hours for 143 epochs with a size of 136.7MB, mAP50 = 0.731, mAP50-95 = 0.458, F1-score = 0.69, inference time while running it on raspberry pi 4 (S1) = 13.510 seconds per frame and inference time while running it on the laptop (S2) = 2.118 seconds per frame. Also, Table 3 shows an F1-score comparison between all models trained for CRDDC2022 [23].

Table 3: Comparison of performance of different YOLO models and number of classes.

| Model | Training Time (hours) | Epochs | Size (MB) | mAP50 | mAP50-90 | F1-Score | Inference Time S1 (sec/frame) | Inference Time S2 (sec/frame) |
|---|---|---|---|---|---|---|---|---|
| YOLOv8s (4 classes) | 10.858 | 187 | 22.5 | 0.597 | 0.315 | 0.59 | 3.054 | 0.376 |
| YOLOv8s (8 classes) | 12.423 | 200 | 22.5 | 0.707 | 0.437 | 0.67 | 3.054 | 0.376 |
| YOLOv8n (4 classes) | 12.987 | 274 | 6.2 | 0.577 | 0.3 | 0.58 | 1.035 | 0.200 |
| YOLOv5n (4 classes) | 14.783 | 300 | 3.9 | 0.54 | 0.249 | 0.56 | 1.074 | 0.231 |
| YOLOv5s (4 classes) | 14.847 | 262 | 14.5 | 0.568 | 0.272 | 0.58 | 3.112 | 0.412 |
| YOLOv8n (8 classes) | 14.911 | 296 | 6.2 | 0.689 | 0.423 | 0.66 | 1.035 | 0.200 |
| YOLOv8m (4 classes) | 15.466 | 160 | 52.0 | 0.615 | 0.326 | 0.60 | 4.717 | 0.737 |
| YOLOv5m (4 classes) | 16.386 | 198 | 42.2 | 0.597 | 0.298 | 0.60 | 4.748 | 0.778 |
| YOLOv8x (8 classes) | 32.526 | 143 | 136.7 | 0.731 | 0.458 | 0.69 | 13.510 | 2.118 |

Table 4: F1-Score of other published systems utilizing YOLO.

| Team Name | Model | F1-Score for 6 countries |
|---|---|---|
| ShiYu_SeaView | Ensemble model based on YOLO-series and Faster RCNN-series models | 0.770 |
| DongjunJeong | YOLOv5x P5 and P6 Ensemble with Image patch | 0.743 |
| MDPT | YOLOv7 with train and test image augmentations, label smoothing, and coordinate attentions | 0.741 |
| IMSC | YOLOv5 based ensemble. | 0.728 |
| IRCV-URV | Yolov7-based ensemble | 0.726 |
| TUT | YPLNet (Yolov5s + Pyramid Squeeze Attention + Largefield Contextual Feature Integration) | 0.694 |
| kubapok | YOLOv5 based ensemble | 0.603 |
| Proposed | YOLOv5n, YOLOv5s and YOLOv5m | 0.56, 0.58, 0.60 |
| Proposed | YOLOv8n, YOLOv8s, YOLOv8m | 0.58, 0.59, 0.60 |

### *d) Results of real-world testing*

A GUI has been built to provide users with friendly access to the developed system such that they can control and manage different functions within the system.

*Figure 4* shows the various buttons to control those functions. The GUI contains two buttons (Mode 1 and Mode 2) to switch between Scenario 1 and Scenario 2, as explained above. The “Select Video” button is used to test a prerecorded video, while the (“Run Object Detection” & ”Stop Detection”) are used to control starting and stopping the detection model. In order to enhance the processing time and count correctly the detected anomalies, a text field has been added to control the frame skipping functionality.

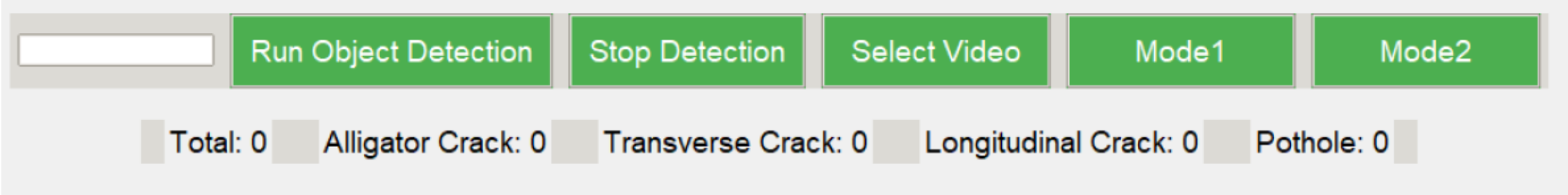

Figure 4: GUI of the developed system, see [24].

B) Experimental setup

The developed system is composed of the following parts:

1- Raspberry pi 4 8 GB to represent the surveillance car and other vehicles.

2- Microsoft Surface6pro with i5-8250U GPU, 8 GB RAM to present the RSU.

3- Logitech C310 camera.

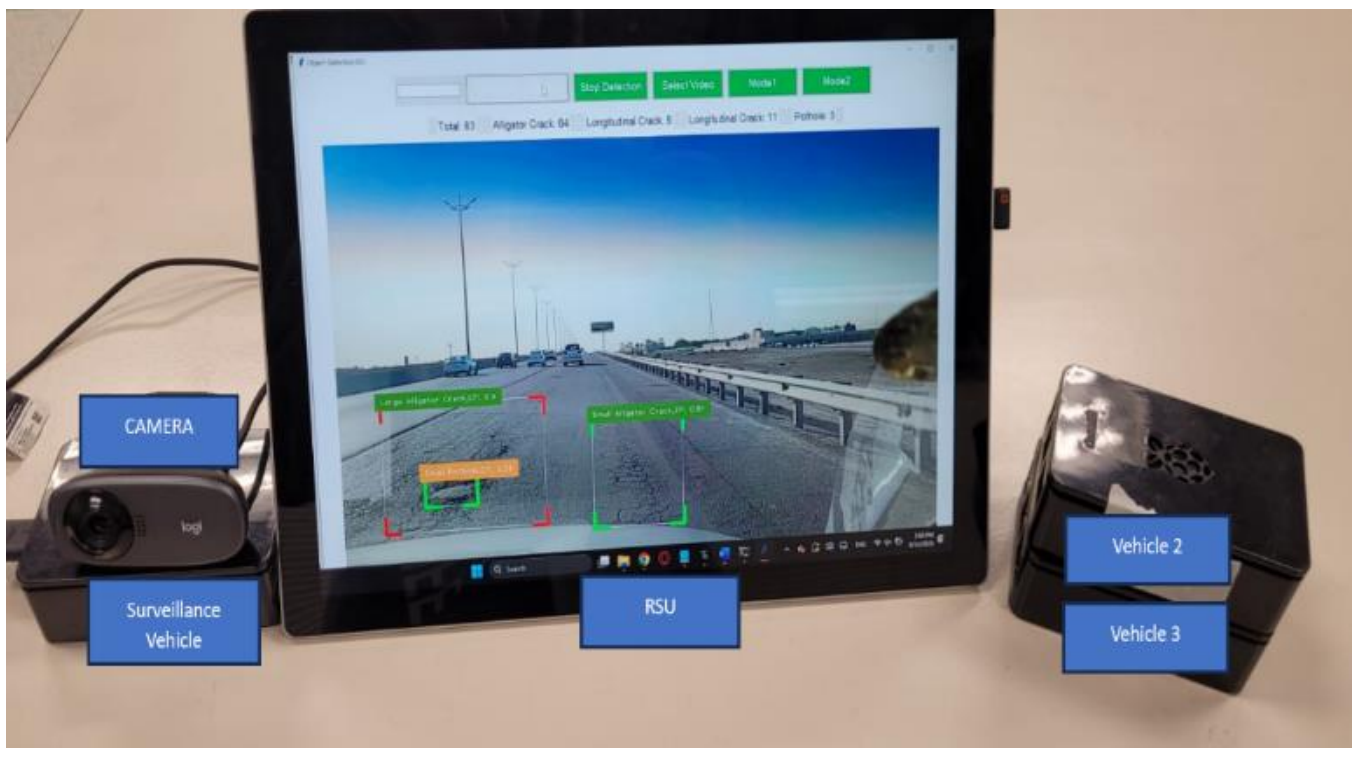

Figure 5: Experimental setup.

The system has been tested in real-world scenarios by capturing offline videos from roads in Saudi Arabia and using them in the system to check its functionality. Figure 6 shows examples of the results detected by our system. The corner indicator changes color from green to red, if the detected anomaly is large to enhance the result readability in real-time. Figure 7 shows the warning notifications sent by RSU to neighboring emulated vehicles. In addition, Figure 8 shows the captured frames of detected anomalies which are sent by the RSU nodes to the cloud side to be stored there for further processing.

Table 5 shows the experimental results for different vehicle's speeds. The yellow marked rows show the optimal settings regarding skipped frames and not producing duplicate instances of anomalies. We can observe that the proposed approach of skipping frames is performing well by minimizing the processing time as well as the number of duplicate frames. However, for the speed of 20 km/h, if we skip 30 frames, there is only one missing anomaly and the processing time is the least. On the other hand, for the speed of 40 km/h and skipping 10 frames, the system captured all anomalies but with 2 duplicates. Moreover, when the speed is 60 km/h and skipping frames is set to 5, the system captured all anomalies but with 5 duplicates. Therefore, the skipping frame mechanism needs to be more fine tuning.

Table 5: Experimental results on different captured videos of real road cases showing YOLOv8n performance on different car speeds. The yellow marked rows show the optimal settings in regard to skipped frames and not producing duplicate instances of anomalies.

| Vehicle Speed | Frames skipped | Actual anomalies | Found anomalies | # of cracks | #of potholes | Video Length (min) | Missing anomalies | Duplicate anomalies | Processing Time (s) |
|---|---|---|---|---|---|---|---|---|---|
| 20 Km/h | 30 | 18 | 17 | 12 | 5 | 1:43 | 1 | 0 | 20.6 |
| 20 Km/h | 20 | 18 | 30 | 25 | 5 | 1:43 | 0 | 12 | 206 |
| 20 Km/h | 10 | 18 | 54 | 46 | 8 | 1:43 | 0 | 36 | 412 |
| 20 Km/h | 5 | 18 | 113 | 95 | 18 | 1:43 | 0 | 95 | 515 |
| 40 Km/h | 30 | 18 | 9 | 6 | 3 | 1:00 | 9 | 0 | 12 |
| 40 Km/h | 20 | 18 | 11 | 8 | 3 | 1:00 | 7 | 0 | 120 |
| 40 Km/h | 10 | 18 | 20 | 16 | 4 | 1:00 | 0 | 2 | 240 |
| 40 Km/h | 5 | 18 | 48 | 41 | 7 | 1:00 | 0 | 30 | 300 |
| 60 Km/h | 30 | 18 | 6 | 4 | 2 | 0:39 | 12 | 0 | 7.8 |
| 60 Km/h | 20 | 18 | 5 | 3 | 2 | 0:39 | 13 | 0 | 78 |
| 60 Km/h | 10 | 18 | 9 | 7 | 2 | 0:39 | 9 | 0 | 156 |
| 60 Km/h | 5 | 18 | 25 | 20 | 5 | 0:39 | 0 | 7 | 195 |

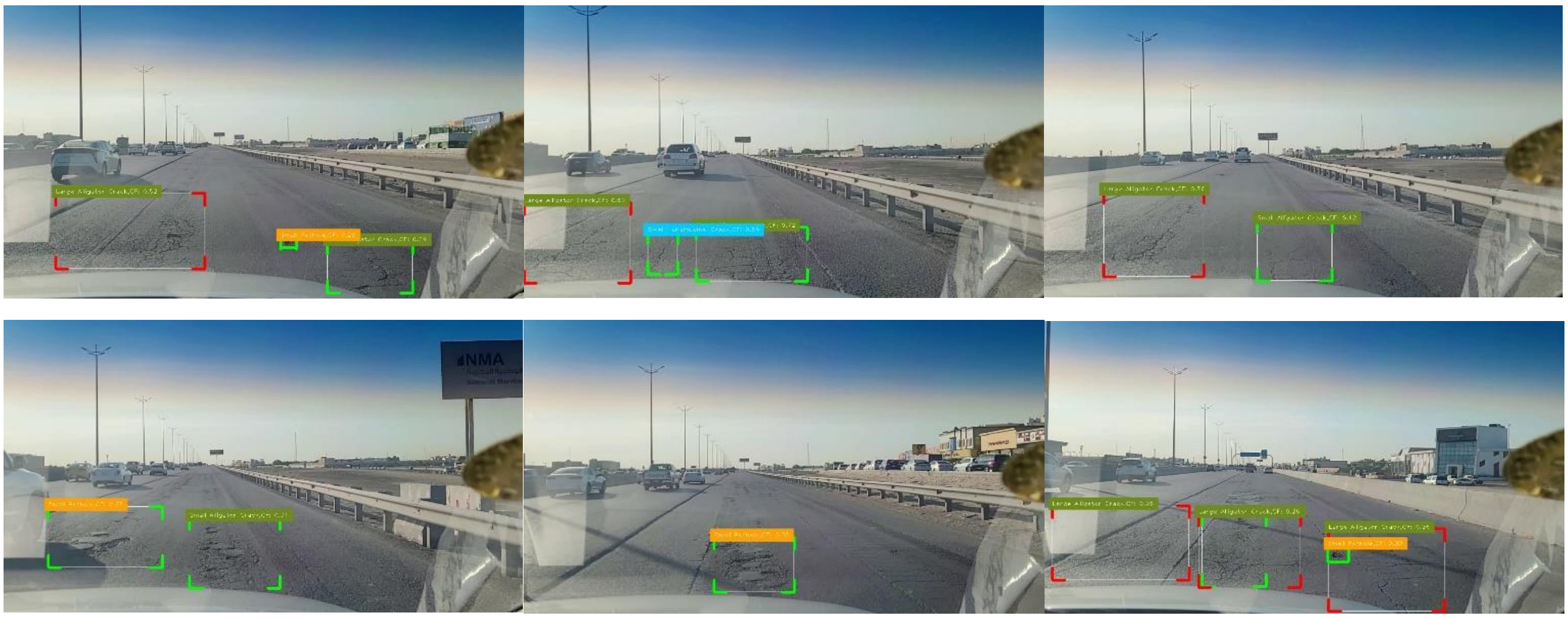

Figure 6: Detected anomalies

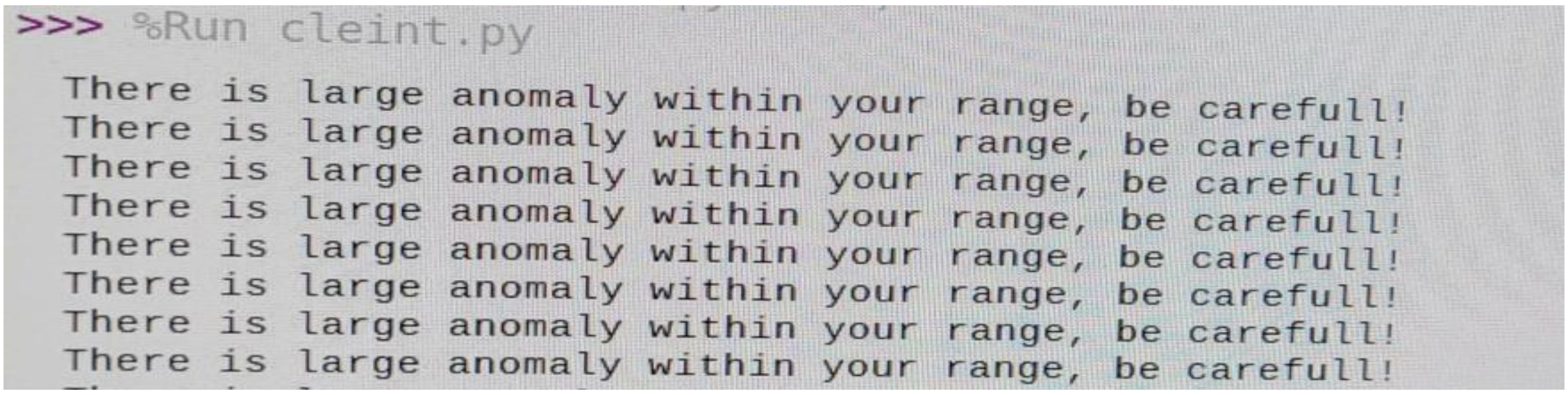

Figure 7: Large anomalies detection alert messages.

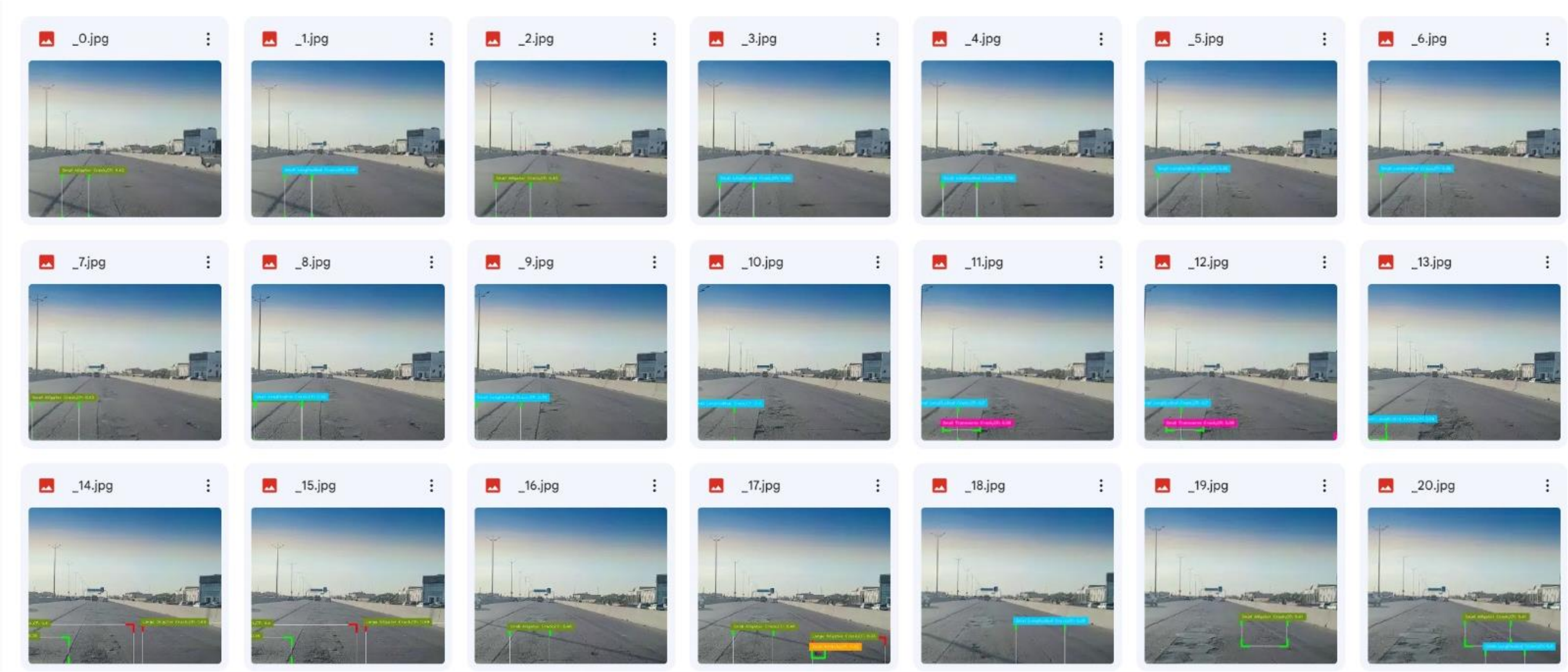

Figure 8: Anomaly detection results in drive.

## V. CONCLUSION AND FUTURE WORK

We hope this study helps in fulfilling partially the objectives of NTS 2030 by enhancing road safety in the Kingdom. The study highlighted the importance of road anomaly detection in the future development of sustainable smart cities. We reviewed different types of methods used in the literature and compared between their differences such as time and space cost. Having gained insights into these methodologies, the study introduced a novel proposed system, which aims to improve on the shortcomings of previous systems. We proposed and tested two modes of operations: Mode-I where the image processing and anomaly detection are performed in the vehicle, while warning messages are sent by RSU, and Mode-II where image capturing is done by the vehicle while, image processing, anomaly detection, and warning messages are all performed by RSU. The developed system was trained on publicly available datasets. Furthermore, the obtained results are compared with contemporary approaches. Finally, the developed system has a GUI which simplifies the user experience and provides multiple features that enhance the system's operation and functionalities.

To improve the system performance, we are planning for the following research steps. Firstly, a GPS module that tracks the route of the car and highlights the location of the anomalies detected previously, which would help the user and authorities to avoid them and fix them. Secondly, when experimenting with new ways to count and detect anomalies, the current system is limited by the fact that anomalies do not have a fixed shape or a fixed set of shapes, which makes them harder for the model to track and keep count of. Further experiments in this area would result in the improvement of both initial detection and later tracking, thus improving the overall accuracy of the model. Another subject that could be explored is extracting more features of detected anomalies such as depth estimation. Capturing this feature would also help in both classification and detection tasks by giving each anomaly more distinct characteristics or features.

## References

[1] The National Transport Strategy 2030, https://rsp.nrsc.gov.sa/api/api/File/get/NTS2030.pdf, accessed October 14, 2023.

[2] P. P.A., N. B. and M. Anuradha, "A Novel Approach for Privacy Preserving In Lane Surface Monitoring Using Fog Computing," *Fifth International Conference on Science Technology Engineering and Mathematics* (ICONSTEM), Chennai, India, 2019, pp. 137-142, doi: 10.1109/ICONSTEM.2019.8918836.

[3] G. Tanganelli, C. Vallati and E. Mingozzi, "A fog-based distributed look-up service for intelligent transportation systems," 2017 IEEE 18th International Symposium on A World of Wireless, Mobile and Multimedia Networks (WoWMoM), Macau, China, 2017, pp. 1-6, doi: 10.1109/WoWMoM.2017.7974357.

[4] S. Singh, R. Chhabra and A. Moudgil, "Classification of Potholes using Convolutional Neural Network Model: A Transfer Learning Approach using Inception ResnetV2," 2nd Edition of IEEE Delhi Section Flagship Conference (DELCON), Rajpura, India, 2023, pp. 1-5, doi: 10.1109/DELCON57910.2023.10127302.

[5] World Health Organization, Global Status Report on Road Safety, WHO, Geneva, Switzerland, 2018.

[6] Y.-M. Kim, Y.-G. Kim, S.-Y. Son, S.-Y. Lim, B.-Y. Choi, and D.-H. Choi, “Review of Recent Automated Pothole-Detection Methods,” Applied Sciences, vol. 12, no. 11, p. 5320, May 2022, doi: https://doi.org/10.3390/app12115320.

[7] N. Ma et al., “Computer Vision for Road Imaging and Pothole Detection: A State-of-the-Art Review of Systems and Algorithms,” arXiv:2204.13590 [cs], Apr. 2022, doi: https://doi.org/10.1093/tse/tdac026.

[8] S. Srivastava, A. Sharma and H. Balot, "Analysis and Improvements on Current Pothole Detection Techniques," 2018 International Conference on Smart Computing and Electronic Enterprise (ICSCEE), Shah Alam, Malaysia, 2018, pp. 1-4, doi: 10.1109/ICSCEE.2018.8538390.

[9] R. Bibi et al., “Edge AI-Based Automated Detection and Classification of Road Anomalies in VANET Using Deep Learning,” Computational Intelligence and Neuroscience, vol. 2021, pp. 1–16, Sep. 2021, doi: https://doi.org/10.1155/2021/6262194.

[10] M. C. Jia Ying, W. Kean Yew, P. J. Yew and A. Yang Her, "Real-time road accident reporting system with location detection using cloud-based data analytics," IECON 2022 – 48th Annual Conference of the IEEE Industrial Electronics Society, Brussels, Belgium, 2022, pp. 1-6, doi: 10.1109/IECON49645.2022.9968507.

[11] M. H. Asad, S. Khaliq, M. H. Yousaf, M. O. Ullah, and A. Ahmad, “Pothole Detection Using Deep Learning: A Real-Time and AI-on-the-Edge Perspective,” Advances in Civil Engineering, vol. 2022, p. e9221211, Apr. 2022, doi: https://doi.org/10.1155/2022/9221211.

[12] M. P. B and S. K.C, “Enhanced pothole detection system using YOLOX algorithm,” *Autonomous Intelligent Systems*, vol. 2, no. 1, Aug. 2022, doi: https://doi.org/10.1007/s43684-022-00037-z.

[13] S.-K. Ryu, T. Kim, and Y.-R. Kim, “Image-Based Pothole Detection System for ITS Service and Road Management System,” Mathematical Problems in Engineering, Sep. 16, 2015. https://www.hindawi.com/journals/mpe/2015/968361/.

[14] D. K. Dewangan and S. P. Sahu, “PotNet: Pothole detection for autonomous vehicle system using convolutional neural network,” Electronics Letters, Dec. 2020, doi: https://doi.org/10.1049/ell2.12062.

[15] J. Terven and D. Cordova-Esparza, “A Comprehensive Review of YOLO: From YOLOv1 to YOLOv8 and Beyond,” arXiv (Cornell University), Apr. 2023, doi: https://doi.org/10.48550/arxiv.2304.00501.

[16] “YOLOv8 vs. YOLOv5: Choosing the Best Object Detection Model,” www.augmentedstartups.com. https://www.augmentedstartups.com/blog/yolov8-vs-yolov5-choosing-the-best-object-detection-model#:~:text=Conclusion- (accessed Aug. 22, 2023).

[17] “YOLOv8 Object Detection Model,” roboflow.com. https://roboflow.com/model/yolov8- (accessed Aug. 22, 2023).

[18] G. Jocher, A. Chaurasia, and J. Qiu, “YOLO by Ultralytics,” GitHub, Jan. 01, 2023. https://github.com/ultralytics/ultralytics

[19] Codrops, “2022 IEEE International Conference on Big Data,” crddc2022.sekilab.global. https://crddc2022.sekilab.global/

[20] G. Jocher, “ultralytics/yolov5,” GitHub, Aug. 21, 2020. https://github.com/ultralytics/yolov5

[21] "RoadDamageDetector," GitHub, Feb. 21, 2023. https://github.com/sekilab/RoadDamageDetector

[22] D. Arya, H. Maeda, S. K. Ghosh, Durga Toshniwal, and Yoshihide Sekimoto, "RDD2022: A multi-national image dataset for automatic Road Damage Detection," Sep. 2022, doi: https://doi.org/10.48550/arxiv.2209.08538.

[23] D. Arya et al., "Crowdsensing-based Road Damage Detection Challenge (CRDDC-2022)," Nov. 2022, doi: https://doi.org/10.48550/arxiv.2211.11362.

[24] https://www.youtube.com/watch?v=1mo7fQg7DP8.